\documentclass [12pt] {article}
\usepackage{amssymb,amsfonts}
\usepackage[cp1251]{inputenc}
\usepackage[T2A]{fontenc}
\usepackage[english]{babel}
\usepackage{graphicx}

\newtheorem{theorem}{Theorem}
\newtheorem{proposition}{Proposition}
\newtheorem{corollary}{Corollary}
\newtheorem{lemma}{Lemma}

\def\argmin{{\rm argmin}}
\def\Fluc{{\rm fluc}}

\begin{document}
\title{Online Learning in Case of Unbounded Losses Using the Follow Perturbed Leader Algorithm\footnote{This paper is an extended version of the ALT
2009 conference paper~\cite{Vyu2009}.}}

\author{Vladimir V. V'yugin\footnote{This research was partially supported by Russian foundation for
fundamental research: 09-07-00180-a and 09-01-00709a.
}\\
\small{Institute for Information Transmission Problems},
\small{Russian Academy of Sciences,}\\
\small{Bol'shoi Karetnyi per. 19, Moscow GSP-4, 127994, Russia}\\
\small{e-mail vyugin@iitp.ru} }

\maketitle

\begin{abstract}
In this paper the sequential prediction problem with expert
advice is considered for the case where losses of experts
suffered at each step cannot be bounded in advance. We present
some modification of Kalai and Vempala algorithm of following
the perturbed leader where weights depend on past losses of the
experts. New notions of a volume and a scaled fluctuation of a
game are introduced. We present a probabilistic algorithm
protected from unrestrictedly large one-step losses. This
algorithm has the optimal performance in the case when the
scaled fluctuations of one-step losses of experts of the pool
tend to zero.

\end{abstract}

\section{Introduction}\label{intr-1}

Experts algorithms are used for online prediction or repeated
decision making or repeated game playing. Starting with the
Weighted Majority Algorithm ({\rm WM}) of Littlestone and
Warmuth~\cite{LiW94} and Vovk's~\cite{Vov90} Aggregating
Algorithm, the theory of Prediction with Expert Advice has
rapidly developed in the recent times. Also, most authors have
concentrated on predicting binary sequences and have used
specific (usually convex) loss functions, like absolute loss,
square and logarithmic loss. A survey can be found in the book
of Lugosi, Cesa-Bianchi~\cite{LCB2007}. Arbitrary losses are
less common, and, as a rule, they are supposed to be bounded in
advance (see well known Hedge Algorithm of Freund and
Shapire~\cite{FrS97}, Normal Hedge~\cite{CFH2009} and other
algorithms).

In this paper, we consider a different general approach --
``Follow the Perturbed Leader -- {\rm FPL}'' algorithm, now
called Hannan's
algorithm~\cite{Han57},~\cite{KaV2003},~\cite{LCB2007}. Under
this approach we only choose the decision that has fared the
best in the past -- the leader. In order to cope with adversary
some randomization is implemented by adding a perturbation to
the total loss prior to selecting the leader. The goal of the
learner's algorithm is to perform almost as well as the best
expert in hindsight in the long run. The resulting {\rm FPL}
algorithm has the same performance guarantees as {\rm WM}-type
algorithms for fixed learning rate and bounded one-step losses,
save for a factor $\sqrt{2}$.

Prediction with Expert Advice considered in this paper proceeds
as follows. We are asked to perform sequential actions at times
$t=1,2,\dots, T$. At each time step $t$, experts $i=1,\dots N$
receive results of their actions in form of their losses
$s_t^i$  - arbitrary real numbers.

At the beginning of the step $t$ {\it Learner}, observing cumulating losses
$s^i_{1:t-1}=s^i_1+\dots +s^i_{t-1}$ of all experts
$i=1,\dots N$, makes a decision to follow one of these experts,
say Expert $i$. At the end of step $t$ {\it Learner} receives
the same loss $s^i_t$ as Expert $i$ at step $t$ and suffers
{\it Learner's} cumulative loss $s_{1:t}=s_{1:t-1}+s^i_t$.

In the traditional framework, we suppose that one-step losses of all experts
are bounded, for example, $0\le s^i_t\le 1$ for all $i$ and $t$.

Well known simple example of a game with two experts shows that Learner
can perform much worse than each expert:
let the current losses of two experts on steps $t=0,1,\dots ,6$ be
$s^1_{0,1,2,3,4,5,6}=(\frac{1}{2},0,1,0,1,0,1)$ and
$s^2_{0.1,2,3,4,5,6}=(0,1,0,1,0,1,0)$.
Evidently, the ``Follow Leader'' algorithm always chooses the wrong prediction.

When the experts one-step losses are bounded, this problem has been solved
using randomization of the experts cumulative losses.
The method of following the perturbed leader was discovered by
Hannan~\cite{Han57}. Kalai and Vempala~\cite{KaV2003} rediscovered
this method and published a simple proof
of the main result of Hannan. They called an algorithm of this type
{\rm FPL} (Following the Perturbed Leader).

The {\rm FPL} algorithm outputs prediction of an expert $i$ which minimizes
$$
s^i_{1:t-1}-\frac{1}{\epsilon}\xi^i,
$$
where $\xi^i$, $i=1,\dots N$, $t=1,2,\dots$,
is a sequence of i.i.d random variables distributed according
to the exponential distribution with the density $p(x)=\exp\{-x\}$,
and $\epsilon$ is {\it a learning rate}.

Kalai and Vempala~\cite{KaV2003} show that the expected cumulative
loss of the {\rm FPL} algorithm has the upper bound
$$
E(s_{1:t})\le (1+\epsilon)\min\limits_{i=1,\dots , N}
s^i_{1:t}+\frac{\log N}{\epsilon},
$$
where $\epsilon$ is a positive real number such that $0<\epsilon<1$ is a
learning rate, $N$ is the number of experts.

Hutter and Poland~\cite{HuP2004}, \cite{HuP2005} presented a
further developments of the {\rm FPL} algorithm for countable
class of experts, arbitrary weights and adaptive learning rate.
Also, {\rm FPL} algorithm is usually considered for bounded
one-step losses: $0\le s^i_t\le 1$ for all $i$ and $t$. Using a
variable learning rate, an optimal upper bound was obtained
in~\cite{HuP2005}~~:
$$
E(s_{1:t})\le\min\limits_{i=1,\dots , N}
s^i_{1:t}+2\sqrt{2T\ln N}.
$$
Most papers on prediction with expert advice either consider
bounded losses or assume the existence of a specific loss
function (see~\cite{LCB2007}). We allow losses at any step to
be unbounded. The notion of a specific loss function is not
used.

The setting allowing unbounded one-step losses do not have wide
coverage in literature; we can only refer reader to
\cite{AAAGO2006}, \cite{CBMS2007}, \cite{PoHP2005}.

Poland and Hutter~\cite{PoHP2005} have studied the games where
one-step losses of all experts at each step $t$ are bounded from above
by an increasing sequence $B_t$ given in advance.
They presented a learning algorithm which is asymptotically consistent
for $B_t=t^{1/16}$.

Allenberg et al.~\cite{AAAGO2006} have
considered polynomially bounded one-step losses
for a modified version of the Littlestone and Warmuth algorithm~\cite{LiW94}
under partial monitoring.
In full information case, their algorithm has the expected regret
$2\sqrt{N\ln N}(T+1)^{\frac{1}{2}(1+a+\beta})$ in the case where
one-step losses of all
experts $i=1,2,\dots N$ at each step $t$ have the bound $(s^i_t)^2\le t^a$,
where $a>0$, and $\beta>0$ is a parameter of the algorithm.
They have proved that this algorithm is Hannan consistent if
$$
\max\limits_{1\le i\le N}\frac{1}{T}\sum\limits_{t=1}^T (s^i_t)^2<cT^a
$$
for all $T$, where $c>0$ and $0<a<1$.

In this paper, we consider also the case where the loss grows
``faster than polynomial, but slower than exponential''. A
motivating example, where losses of the experts cannot be
bounded in advance, is given in Section~\ref{arbitr-1}.

We present some modification of Kalai and
Vempala~\cite{KaV2003} algorithm of following the perturbed
leader ({\rm FPL}) for the case of unrestrictedly large
one-step expert losses $s^i_t$ not bounded in advance: $s^i_t\in (-\infty,+\infty)$. This
algorithm uses adaptive weights depending on past cumulative
losses of the experts.

The full information case is considered in this paper. We
analyze the asymptotic consistency of our algorithms using
nonstandard scaling. We introduce new notions of {\it the
volume of a game} $v_t=v_{0}+\sum\limits_{j=1}^t\max_i |s^i_j|$
and {\it the scaled fluctuation} of the game $\Fluc(t)=\Delta
v_t/v_t$, where $\Delta v_t=v_t-v_{t-1}$ and $v_{0}$ is a
nonnegative constant.

We show in Theorem~\ref{cor-2} that the algorithm of following the perturbed
leader with adaptive weights constructed in Section~\ref{pos-gain} is
asymptotically consistent in the mean in the case
where $v_t\to\infty$ and $\Delta v_t=o(v_t)$ as $t\to\infty$
with a computable bound. Specifically,
if $\Fluc(t)\le\gamma(t)$ for all $t$, where $\gamma(t)$
is a computable function such that $\gamma(t)=o(1)$ as
$t\to\infty$, our algorithm has the expected regret
$$
2\sqrt{(6+\epsilon)(1+\ln N)}\sum_{t=1}^T(\gamma(t))^{1/2}\Delta v_t,
$$
where $\epsilon>0$ is a parameter of the algorithm.

In case where all losses are nonnegative: $s^i_t\in [0,+\infty)$,
we obtain a regret
\begin{eqnarray*}
2\sqrt{(2+\epsilon)(1+\ln N)}
\sum_{t=1}^T (\gamma(t))^{1/2}\Delta v_t.
\end{eqnarray*}

In particular, this algorithm is asymptotically consistent (in the mean)
in a modified sense
\begin{eqnarray}
\limsup\limits_{T\to\infty}\frac{1}{v_T}
E(s_{1:T}-\min\limits_{i=1,\dots N} s^i_{1:T})\le 0,
\label{vol-1}
\end{eqnarray}
where $s_{1:T}$ is the total loss of our algorithm on steps $1,2,\dots T$,
and $E(s_{1:T})$ is its expectation.

Proposition~\ref{cor-2-1} of Section~\ref{pos-gain1} shows that
if the condition $\Delta v_t=o(v_t)$ is violated the cumulative
loss of any probabilistic prediction algorithm can be much more
than the loss of the best expert of the pool.

In Section~\ref{pos-gain} we present some sufficient conditions under which
our learning algorithm is Hannan consistent.
\footnote
{
This means that (\ref{vol-1}) holds with probability 1, where $E$
is omitted.
}

In particular case, Corollary~\ref{cor-1gg} of Theorem~\ref{cor-2} says
that our algorithm is asymptotically consistent (in the modified sense)
in the case when one-step losses of all experts at each step $t$ are bounded
by $t^a$, where $a$ is a positive real number. We prove this result
under an extra assumption that the volume of the game grows slowly,
$
\liminf\limits_{t\to\infty}v_t/t^{a+\delta}>0,
$
where $\delta>0$ is arbitrary.
Corollary~\ref{cor-1gg} shows that our algorithm is also Hannan
consistent when $\delta>\frac{1}{2}$.

At the end of Section~\ref{pos-gain} we consider some
applications of our algorithm for the case of standard
time-scaling.

In Section~\ref{arbitr-1} we consider an application of our
algorithm for constructing an arbitrage strategy in some game
of buying and selling shares of some stock on financial market.
We analyze this game in the decision theoretic online learning
(DTOL) framework~\cite{FrS97}. We introduce {\it Learner} that
computes weighted average of different strategies with
unbounded gains and losses. To change from the follow leader
framework to DTOL we derandomize our {\rm FPL} algorithm.

\section{Games of prediction with expert advice with
unbounded one-step losses}
\label{pos-gain1}

We consider a game of prediction with expert advice with
arbitrary unbounded one-step losses. At each step $t$ of the
game, all $N$ experts receive one-step losses $s^i_t\in
(-\infty,+\infty)$, $i=1,\dots N$, and the cumulative loss of
the $i$th expert after step $t$ is equal to
$$
s^i_{1:t}=s^i_{1:t-1}+s^i_t.
$$
A probabilistic learning algorithm of choosing an expert outputs at any
step $t$ the probabilities $P\{I_t=i\}$ of following the $i$th expert
given the cumulative losses $s^i_{1:t-1}$
of the experts $i=1,\dots N$ in hindsight.

{\bf Probabilistic algorithm of choosing an expert}.

\noindent FOR $t=1,\dots T$

Given past cumulative losses of the experts $s^i_{1:t-1}$, $i=1,\dots N$,
choose an expert $i$ with probability $P\{I_t=i\}$.

Receive the one-step losses at step $t$ of the expert $s^i_t$
and suffer one-step loss $s_t=s^i_t$ of the master algorithm.

\noindent ENDFOR

The performance of this probabilistic algorithm is measured
in its {\it expected regret}
\begin{eqnarray*}
E(s_{1:T}-\min\limits_{i=1,\dots N} s^i_{1:T}),
\label{consist-1hh}
\end{eqnarray*}
where the random variable $s_{1:T}$ is the cumulative loss of
the master algorithm, $s^i_{1:T}$, $i=1,\dots N$, are the
cumulative losses of the experts algorithms and $E$ is the
mathematical expectation (with respect to the probability
distribution generated by probabilities $P\{I_t=i\}$,
$i=1,\dots N$, on the first $T$ steps of the game).

In the case of bounded one-step expert losses, $s^i_t\in [0,1]$, and a convex
loss function, the well-known learning algorithms have expected regret
$O(\sqrt{T\log N})$ (see Lugosi, Cesa-Bianchi~\cite{LCB2007}).

A probabilistic algorithm is called {\it asymptotically consistent}
in the mean if
\begin{eqnarray}
\limsup\limits_{T\to\infty}
\frac{1}{T}E(s_{1:T}-\min\limits_{i=1,\dots N} s^i_{1:T})\le 0.
\label{consist-1d}
\end{eqnarray}
A probabilistic learning algorithm is called {\it Hannan consistent} if
\begin{eqnarray}
\limsup\limits_{T\to\infty}
\frac{1}{T}\left(s_{1:T}-\min\limits_{i=1,\dots N} s^i_{1:T}\right)\le 0
\label{consist-1han}
\end{eqnarray}
almost surely, where $s_{1:T}$ is its random cumulative loss.

In this section we study the asymptotical consistency of probabilistic
learning algorithms in the case of unbounded one-step losses.

Notice that when $0\le s^i_t\le 1$ all expert algorithms have total
loss $\le T$ on first $T$ steps.
This is not true for the unbounded case, and there are no reasons to divide
the expected regret (\ref{consist-1d}) by $T$. We change the standard
time scaling (\ref{consist-1d}) and (\ref{consist-1han}) on a new scaling
based on a new notion of volume of a game. We modify the definition
(\ref{consist-1d}) of the normalized expected regret as follows.
Define {\it the volume} of a game at step $t$
$$
v_t=v_{0}+\sum\limits_{j=1}^t\max_i |s^i_j|,
$$
where $v_{0}$ is a nonnegative constant. Evidently, $v_{t-1}\le
v_t$ for all $t$.

A probabilistic learning algorithm is called
{\it asymptotically consistent} in the mean (in the modified sense)
in a game with $N$ experts if
\begin{eqnarray}
\limsup\limits_{T\to\infty}
\frac{1}{v_T}E(s_{1:T}-\min\limits_{i=1,\dots N}s^i_{1:T})\le 0.
\label{mod-consist-1}
\end{eqnarray}
A probabilistic algorithm is called Hannan consistent
(in the modified sense) if
\begin{eqnarray}
\limsup\limits_{T\to\infty}
\frac{1}{v_T}\left(s_{1:T}-\min\limits_{i=1,\dots N} s^i_{1:T}\right)\le 0
\label{consist-1han-1}
\end{eqnarray}
almost surely.

Notice that the notions of asymptotic consistency in the mean
and Hannan consistency may be non-equivalent for unbounded one-step losses.

A game is called {\it non-degenerate} if $v_t\to\infty$ as
$t\to\infty$.

Denote $\Delta v_t=v_t-v_{t-1}$. The number
\begin{eqnarray}\label{dev-1}
{\Fluc}(t)=\frac{\Delta v_t}{v_t}=\frac{\max_i|s^i_t|}{v_t},
\end{eqnarray}
is called {\it scaled fluctuation} of the game at the step $t$.

By definition $0\le {\Fluc}(t)\le 1$ for all $t$ (put $0/0=0$).

The following simple proposition shows that each probabilistic
learning algorithm is not asymptotically optimal in some game
such that $\Fluc(t)\not\to 0$ as $t\to\infty$. For simplicity,
we consider the case of two experts and nonnegative losses.

\begin{proposition}\label{cor-2-1}
For any probabilistic algorithm of choosing an expert and
for any $\epsilon$ such that $0<\epsilon<1$ two experts exist such that
$v_t\to\infty$ as $t\to\infty$ and
\begin{eqnarray*}
\Fluc(t)\ge 1-\epsilon,
\nonumber\\
\frac{1}{v_t}E(s_{1:t}-\min\limits_{i=1,2} s^i_{1:t})\ge\frac{1}{2}(1-\epsilon)
\end{eqnarray*}
for all $t$.
\end{proposition}
{\it Proof}.
Given a probabilistic algorithm of choosing an expert and $\epsilon$
such that $0<\epsilon<1$, define recursively one-step losses $s^1_t$
and $s^2_t$ of expert 1 and expert 2 at any step $t=1,2,\dots$ as follows.
By $s^1_{1:t}$ and $s^2_{1:t}$ denote the cumulative losses of these experts
incurred at steps $\le t$, let $v_t$ be the corresponding volume,
where $t=1,2,\dots$.

Define $v_0=1$ and $M_t=4v_{t-1}/\epsilon$ for all $t\ge 1$.
For $t\ge 1$, define $s^1_t=0$ and $s^2_t=M_t$ if $P\{I_t=1\}\ge\frac{1}{2}$,
and define $s^1_t=M_t$ and $s^2_t=0$ otherwise.

Let $s_t$ be one-step loss of the master algorithm and $s_{1:t}$
be its cumulative loss at step $t\ge 1$. We have
$$
E(s_{1:t})\ge E(s_t)=s^1_tP\{I_t=1\}+s^2_tP\{I_t=2\}\ge\frac{1}{2}M_t
$$
for all $t\ge 1$.
Also, since $v_t=v_{t-1}+M_t=(1+4/\epsilon)v_{t-1}$
and $\min\limits_i s^i_{1:t}\le v_{t-1}$,
the normalized expected regret of the
master algorithm is bounded from below
\begin{eqnarray*}
\frac{1}{v_t}
E(s_{1:t}-\min\limits_i s^i_{1:t})\ge
\frac{2/\epsilon-1}{1+4/\epsilon}\ge\frac{1}{2}(1-\epsilon).
\label{mod-consist-1a}
\end{eqnarray*}
for all $t$. By definition
$$
\Fluc(t)=\frac{M_t}{v_{t-1}+M_t}=\frac{1}{1+\epsilon/4}\ge 1-\epsilon
$$
for all $t$.
$\triangle$

Proposition~\ref{cor-2-1} shows that we should impose some
restrictions of asymptotic behavior of $\Fluc(t)$ to prove the
asymptotic consistency of a probabilistic algorithm.

\section{The Follow Perturbed Leader algorithm with adaptive weights}
\label{pos-gain}

In this section we construct the {\rm FPL} algorithm with
adaptive weights protected from unbounded one-step losses.

Let $\gamma(t)$ be a computable non-increasing real function
such that $0<\gamma(t)<1$ for all $t$ and $\gamma(t)\to 0$ as
$t\to\infty$; for example, $\gamma(t)=1/t^\delta$, where
$\delta>0$. Let also $a$ be a positive real number. Define
\begin{eqnarray}
\alpha_t=\frac{1}{2}\left(1-\frac
{\ln\frac{a(1+\ln N)}{2(e^{3/a}-1)}}{\ln\gamma(t)}\right) \mbox{ and }
\label{optalpha-1}
\\
\mu_t=a(\gamma(t))^{\alpha_t}=\sqrt{\frac{2a(e^{3/a}-1)}{(1+\ln N)}}(\gamma(t))^{1/2}
\label{mu-1}
\end{eqnarray}
for all $t$, where $e=2.72\dots$ is the base of the natural
logarithm. \footnote { The choice of the optimal value of
$\alpha_t$ will be explained later. It will be obtained by
minimization of the corresponding member of the sum
(\ref{j_mu-1af}).
}

Without loss of generality we suppose that $
\gamma(t)<\min\{A,A^{-1}\} $ for all $t$, where
$$A=\frac{2(e^{3/a}-1)}{a(1+\ln N)}.$$ We can obtain this
choosing an appropriate value of the initial constant $v_{0}$.
Then $0<\alpha_t<1$ for all $t$.

We consider an {\rm FPL} algorithm with a variable learning rate
\begin{eqnarray}\label{eps-1}
\epsilon_t=\frac{1}{\mu_tv_{t-1}},
\end{eqnarray}
where $\mu_t$ is defined by (\ref{mu-1})
and the volume $v_{t-1}$ depends on experts actions on steps $<t$.
By definition $v_t\ge v_{t-1}$ and $\mu_t\le\mu_{t-1}$ for $t=1,2,\dots$.
Also, by definition $\mu_t\to 0$ as $t\to\infty$.

Let $\xi_t^1$,\dots $\xi_t^N$, $t=1,2,\dots$,
be a sequence of i.i.d random variables
distributed according to the density $p(x)=\exp\{-x\}$.
In what follows we omit the lower index $t$.

We suppose without loss of generality that $s^i_0=v_0=0$
for all $i$ and $\epsilon_0=\infty$.

The {\rm FPL} algorithm is defined as follows:

{\bf FPL algorithm PROT}.

\noindent FOR $t=1,\dots T$

Choose an expert with the minimal perturbed cumulated loss on steps
$<t$
\begin{eqnarray}\label{m-1}
I_t=\argmin_{i=1,2,\dots N}\{s^i_{1:t-1}-\frac{1}{\epsilon_t}\xi^i\}.
\end{eqnarray}
Receive one-step losses $s_t^i$ for experts $i=1,\dots , N$,
define $v_{t}=v_{t-1}+\max\limits_{i}s^{i}_{t}$ and
$\epsilon_{t+1}$ by (\ref{eps-1}).

Receive one-step loss $s_{t}=s^{I_t}_t$ of the master
algorithm.

\noindent ENDFOR

Let $s_{1:T}=\sum\limits_{t=1}^T s^{I_t}_t$ be the cumulative loss
of the {\rm FPL} algorithm on steps $\le T$.

The following theorem shows that if the game is non-degenerate and
$\Delta v_t=o(v_t)$ as $t\to\infty$
with a computable bound then the {\rm FPL}-algorithm with variable
learning rate (\ref{eps-1}) is asymptotically consistent.

We suppose that the experts are oblivious, i.e., they do not
use in their work random actions of the learning algorithm. The
inequality (\ref{main-refret-1}) of Theorem~\ref{cor-2} below
is reformulated and proved for non-oblivious experts at the end
this section.

\begin{theorem}\label{cor-2}
Let $\gamma(t)$ be a computable non-increasing real function
such that $0\le\gamma(t)\le 1$ and
\begin{eqnarray}\label{bound-fluc-1}
\Fluc(t)\le\gamma(t)
\end{eqnarray}
for all $t$. Then for any $\epsilon>0$ the expected cumulated loss of
the {\rm FPL} algorithm {\rm PROT} with variable learning rate (\ref{eps-1}),
where parameter $a$ depends on $\epsilon$, is bounded:
\begin{eqnarray}
E(s_{1:T})\le\min_i s^i_{1:T}+2\sqrt{(6+\epsilon)(1+\ln N)}
\sum_{t=1}^T (\gamma(t))^{1/2}\Delta v_t
\label{main-refret-1}
\end{eqnarray}
for all $t$.

In case of nonnegative unbounded losses $s_t^i\in [0,+\infty)$ we have a bound
\begin{eqnarray}
E(s_{1:T})\le\min_i s^i_{1:T}+2\sqrt{(2+\epsilon)(1+\ln N)}
\sum_{t=1}^T (\gamma(t))^{1/2}\Delta v_t.
\label{main-refret-1kk}
\end{eqnarray}

Let also,
the game be non-degenerate and $\gamma(t)\to 0$ as
$t\to\infty$. Then the algorithm {\rm PROT} is asymptotically
consistent in the mean
\begin{eqnarray}
\limsup\limits_{T\to\infty}\frac{1}{v_T}E(s_{1:T}-\min\limits_{i=1,\dots N}
s^i_{1:T})\le 0.
\label{mod-consist-1hhh}
\end{eqnarray}
\end{theorem}
{\it Proof}. The proof of this theorem follows the
proof-scheme of~\cite{HuP2004} and~\cite{KaV2003}.

Let $\alpha_t$ be a sequence of real numbers defined by (\ref{optalpha-1});
recall that $0<\alpha_t<1$ for all $t$.

The analysis of optimality
of the FPL algorithm is based on an intermediate predictor {\rm IFPL}
(Infeasible {\rm FPL}) with the learning rate $\epsilon'_t$ defined by
(\ref{eps-2}).

{\bf IFPL algorithm}.

\noindent FOR $t=1,\dots T$

Define the learning rate
\begin{eqnarray}\label{eps-2}
\epsilon'_t=\frac{1}{\mu_t v_t}, \mbox{ where } \mu_t=a(\gamma(t))^{\alpha_t},
\end{eqnarray}
$v_t$ is the volume of the game at step $t$ and $\alpha_t$
is defined by (\ref{optalpha-1}).

Choose an expert with the minimal perturbed cumulated loss on steps
$\le t$
\begin{eqnarray*}
J_t=\argmin_{i=1,2,\dots N}\{s^i_{1:t}-\frac{1}{\epsilon'_t}\xi^i\}.
\end{eqnarray*}
Receive the one step loss $s^{J_t}_t$ of the {\rm IFPL} algorithm.

\noindent ENDFOR

The {\rm IFPL} algorithm predicts under the knowledge of
$s^i_{1:t}$, $i=1,\dots N$ (and $v_t$),
which may not be available at beginning of step $t$. Using unknown value of
$\epsilon'_t$ is the main distinctive feature of our version of {\rm IFPL}.

For any $t$, we have
$
I_t=\argmin_i\{s_{1:t-1}^i-\frac{1}{\epsilon_t}\xi^i\}$ and
$J_t=\argmin_i\{s_{1:t}^i-\frac{1}{\epsilon'_t}\xi^i\}=
\argmin_i\{s_{1:t-1}^i+s_t^i-\frac{1}{\epsilon'_t}\xi^i\}$.

The expected one-step and cumulated losses of the {\rm FPL} and
{\rm IFPL} algorithms at steps $t$ and $T$ are denoted
\begin{eqnarray*}
l_t=E(s_t^{I_t}) \mbox{ and } r_t=E(s_t^{J_t}),
\nonumber
\\
l_{1:T}=\sum\limits_{t=1}^T l_t \mbox{ and } r_{1:T}=\sum\limits_{t=1}^T r_t,
\end{eqnarray*}
respectively, where $s_t^{I_t}$ is the one-step loss of the {\rm FPL} algorithm
at step $t$ and $s_t^{J_t}$ is the one-step loss of the {\rm IFPL} algorithm,
and $E$ denotes the mathematical expectation.

\begin{lemma}\label{fpl-ifpl-1}
The cumulated expected losses of the {\rm FPL} and {\rm IFPL} algorithms
with rearning rates defined by (\ref{eps-1}) and (\ref{eps-2})
satisfy the inequality
\begin{eqnarray}
l_{1:T}\le r_{1:T}+2(e^{3/a}-1)
\sum\limits_{t=1}^T(\gamma(t))^{1-\alpha_t}\Delta v_t
\label{thh-1a}
\end{eqnarray}
for all $T$, where $\alpha_t$ is defined by (\ref{optalpha-1}).
\end{lemma}
{\it Proof}.
Let $c_1,\dots c_N$ be nonnegative real numbers and
\begin{eqnarray*}
m_j=\min\limits_{i\not =j}\{s_{1:t-1}^i-\frac{1}{\epsilon_t}c_i\},
\nonumber
\\
m'_j=\min\limits_{i\not =j}\{s_{1:t}^i-\frac{1}{\epsilon'_t}c_i\}=
\min\limits_{i\not =j}\{s_{1:t-1}^i+s_t^i-\frac{1}{\epsilon'_t}c_i\}.
\end{eqnarray*}

Let $m_j=s^{j_1}_{1:t-1}-\frac{1}{\epsilon_t}c_{j^1}$ and
$m'_j=s_{1:t}^{j_2}-\frac{1}{\epsilon'_t}c_{j_2}=
s_{1:t-1}^{j_2}+s^{j_2}_t-\frac{1}{\epsilon'_t}c_{j_2}$.
By definition and since $j_2\not =j$ we have
\begin{eqnarray}
m_j=s_{1:t-1}^{j_1}-\frac{1}{\epsilon_t}c_{j_1}\le
s^{j_2}_{1:t-1}-\frac{1}{\epsilon_t}c_{j^2}\le
s^{j_2}_{1:t-1}+s^{j_2}_t-\frac{1}{\epsilon_t}c_{j_2}=
\label{m'-m-1f}
\\
s^{j_2}_{1:t}-\frac{1}{\epsilon'_t}c_{j_2}+
\left(\frac{1}{\epsilon'_t}-\frac{1}{\epsilon_t}\right)c_{j_2}=
m'_j+\left(\frac{1}{\epsilon'_t}-\frac{1}{\epsilon_t}\right)c_{j_2}.
\label{m'-m-1}
\end{eqnarray}

We compare conditional probabilities
$P\{I_t=j|\xi^i=c_i, i\not =j\}$ and $P\{J_t=j|\xi^i=c_i, i\not =j\}$.

The following chain of equalities and inequalities is valid:
\begin{eqnarray}
P\{I_t=j|\xi^i=c_i, i\not =j\}=
~~~~~~\nonumber\\
P\{s^j_{1:t-1}-\frac{1}{\epsilon_t}\xi^j
\le m_j|\xi^i=c_i, i\not =j\}=~~~~~~
\nonumber
\\
P\{\xi^j\ge\epsilon_t(s^j_{1:t-1}-m_j)
|\xi^i=c_i, i\not =j\}=~~~~~~
\nonumber
\\
P\{\xi^j\ge\epsilon'_t(s^j_{1:t-1}-m_j)+
(\epsilon_t-\epsilon'_t)(s^j_{1:t-1}-m_j)|\xi^i=c_i, i\not =j\}\le~~~~~~
\label{iineq-1}
\\
P\{\xi^j\ge\epsilon'_t(s^j_{1:t-1}-m_j)+~~~~~~~~~~~
\nonumber
\\
(\epsilon_t-\epsilon'_t)
(s^j_{1:t-1}-s_{1:t-1}^{j_2}+
\frac{1}{\epsilon_t}c_{j_2})
|\xi^i=c_i, i\not =j\}=~~~~~~
\label{1-1iq}
\\
\exp\{-(\epsilon_t-\epsilon'_t)(s_{1:t-1}^j-s^{j_2}_{1:t-1})\}\times~~~~~~~~~~~
\label{1-1q}
\\
P\{\xi^j\ge\epsilon'_t(s^j_{1:t-1}-m_j)+
(\epsilon_t-\epsilon'_t)\frac{1}{\epsilon_t}
c_{j_2}|\xi^i=c_i, i\not =j\}\le~~~~~~
\label{m'-m-1a}
\\
\exp\{-(\epsilon_t-\epsilon'_t)(s_{1:t-1}^j-s^{j_2}_{1:t-1})\}\times~~~~~~~~~~~
\nonumber
\\
P\{\xi^j\ge
\epsilon'_t(s^j_{1:t}-s_t^j-m'_j-\left(\frac{1}{\epsilon'_t}-
\frac{1}{\epsilon_t}\right)c_{j_2})+~~~~~~~~~~~
\label{m'-m-1b} 
\\
(\epsilon_t-\epsilon'_t)\frac{1}{\epsilon_t}c_{j_2}|
\xi^i=c_i, i\not =j\}=~~~~~~
\label{1-2iq}
\\
\exp\{-(\epsilon_t-\epsilon'_t)(s_{1:t-1}^j-
s^{j_2}_{1:t-1})+\epsilon'_t s^j_t\}\times~~~~~~~~~~~
\label{1-2q}\\
P\{\xi^j\ge\epsilon'_t(s^j_{1:t}-m'_j)|\xi^i=c_i, i\not =j\}=~~~~~~
\nonumber
\\
\exp\left\{-\left(\frac{1}{\mu_t v_{t-1}}-\frac{1}{\mu_t v_t}\right)
(s_{1:t-1}^j-s^{j_2}_{1:t-1})+\frac{s^j_t}{\mu_t v_t}\right\}\times~~~~~~~~~~~
\label{1-2}
\\
P\{\xi^j>\frac{1}{\mu_t v_t}(s^j_{1:t}-m'_j)|\xi^i=c_i, i\not =j\}
\le~~~~~~~~
\nonumber
\\
\exp\left\{-\frac{\Delta v_t}
{\mu_tv_t} \frac{(s_{1:t-1}^j-s^{j_2}_{1:t-1})}{v_{t-1}}+
\frac{\Delta v_t}{\mu_t v_t}\right\}\times~~~~~~~~~~~
\label{gamma-1}
\\
P\{\xi^j>\frac{1}{\mu_t v_t}(s^j_{1:t}-m'_j)|\xi^i=c_i, i\not =j\}=~~~~~~
\nonumber
\\
\exp\left\{\frac{\Delta v_t}{\mu_t v_t}
\left(1-\frac{s_{1:t-1}^j-
s^{j_2}_{1:t-1}}{v_{t-1}}\right)\right\}
P\{J_t=1|\xi^i=c_i, i\not =j\}.~~~~~~
\label{1-3}
\end{eqnarray}
Here the inequality (\ref{iineq-1})-(\ref{1-1iq}) follows from
(\ref{m'-m-1f}) and $\epsilon_t\ge\epsilon'_t$. We have used
twice, in change from (\ref{1-1iq}) to (\ref{1-1q}) and in
change from (\ref{1-2iq}) to (\ref{1-2q}), the equality
$P\{\xi>a+b\}=e^{-b}P\{\xi>a\}$ for any random variable $\xi$
distributed according to the exponential law. The equality
(\ref{m'-m-1a})-(\ref{m'-m-1b}) follows from (\ref{m'-m-1}). We
have used in change from (\ref{1-2}) to (\ref{gamma-1}) the
equality $v_t-v_{t-1}=\Delta v_t$ and the inequality
$|s^j_t|\le\Delta v_t$ for all $j$ and $t$.

The ratio in the exponent (\ref{1-3}) is bounded~:
\begin{eqnarray}\label{ij-1}
\left|\frac{s_{1:t-1}^j-s^{j_2}_{1:t-1}}{v_{t-1}}\right|\le 2,
\end{eqnarray}
since $\left|\frac{s^i_{1:t-1}}{v_{t-1}}\right|\le 1$ for all
$t$ and $i$.

Therefore, we obtain
\begin{eqnarray}
P\{I_t=j|\xi^i=c_i, i\not =j\}\le
\nonumber
\\
\exp\left\{\frac{3}{\mu_t}\frac{\Delta v_t}
{v_t}\right\}P\{J_t=j|\xi^i=c_i, i\not =j\}\le
\nonumber
\\
\exp\{(3/a)(\gamma(t))^{1-\alpha_t}\}P\{J_t=j|\xi^i=c_i, i\not =j\}.
\label{bo-1}
\end{eqnarray}

Since, the inequality (\ref{bo-1}) holds for all $c_i$,
it also holds unconditionally
\begin{eqnarray}
P\{I_t=j\}\le \exp\{(3/a)(\gamma(t))^{1-\alpha_t}\}P\{J_t=j\}.
\label{i-00}
\end{eqnarray}
for all $t=1,2,\dots$ and $j=1,\dots N$.

Since $s^j_t+\Delta v_t\ge 0$ for all $j$ and $t$, we obtain
from (\ref{i-00})
\begin{eqnarray}
l_t+\Delta v_t=E(s^{I_t}_t+\Delta v_t)=\sum\limits_{j=1}^N
(s^j_t+\Delta v_t)P(I_t=j)\le
\nonumber\\
\exp\{(3/a)(\gamma(t))^{1-\alpha_t}\}\sum\limits_{j=1}^N
(s^j_t+\Delta v_t) P(J_t=j)=
\nonumber\\
\exp\{(3/a)(\gamma(t))^{1-\alpha_t}\}(E(s^{J_t}_t)+\Delta v_t)=
\nonumber\\
\exp\{(3/a)(\gamma(t))^{1-\alpha_t}\}(r_t+\Delta v_t)\le
\nonumber\\
(1+(e^{3/a}-1))(\gamma(t))^{1-\alpha_t})(r_t+\Delta v_t)=
\nonumber\\
r_t+\Delta v_t+(e^{3/a}-1)(\gamma(t))^{1-\alpha_t}(r_t+\Delta v_t)\le
\nonumber\\
r_t+\Delta v_t+2(e^{3/a}-1)(\gamma(t))^{1-\alpha_t}\Delta v_t.
\label{expect-1}
\end{eqnarray}
In the last line of (\ref{expect-1}) we have used the
inequality $|r_t|\le\Delta v_t$ for all $t$ and the inequality
$\exp\{3r\}\le 1+(e^{3}-1)r$ for all $0\le r\le 1$.

Subtracting $\Delta v_t$ from both sides of the inequality
(\ref{expect-1}) and summing it by $t=1,\dots T$, we obtain
\begin{eqnarray*}
l_{1:T}\le r_{1:T}+2(e^{3/a}-1)\sum_{t=1}^T (\gamma(t))^{1-\alpha_t}\Delta v_t
\end{eqnarray*}
for all $T$. Lemma~\ref{fpl-ifpl-1} is proved. $\triangle$

The following lemma, which is an analogue of the result
from~\cite{KaV2003}, gives a bound for the IFPL algorithm.
\begin{lemma}\label{IFPL-1}
The expected cumulative loss of the {\rm IFPL} algorithm with the
learning rate (\ref{eps-2}) is bounded~:
\begin{eqnarray}
r_{1:T}\le\min_i s^i_{1:T}+a(1+\ln N)
\sum_{t=1}^T(\gamma(t))^{\alpha_t}\Delta v_t
\label{ii-ff}
\end{eqnarray}
for all $T$, where $\alpha_t$ is defined by (\ref{optalpha-1}).
\end{lemma}
{\it Proof}. The proof is along the line of the proof from Hutter and
Poland~\cite{HuP2004} with an exception that now the sequence $\epsilon'_t$
is not monotonic.

Let in this proof, ${\bf s_t}=(s^1_t,\dots s^N_t)$ be a vector of one-step losses
and ${\bf s_{1:t}}=(s^1_{1:t},\dots s^N_{1:t})$ be a vector of cumulative losses
of the experts algorithms.
Also, let ${\bf \xi}=(\xi^1,\dots \xi^N)$ be a vector whose coordinates are
random variables.

Recall that $\epsilon'_t=1/(\mu_t v_t)$, $\mu_t\le\mu_{t-1}$
for all $t$, and $v_0=0$, $\epsilon'_0=\infty$.

Define
$
{\bf \tilde s_{1:t}}={\bf s_{1:t}}-\frac{1}{\epsilon'_t}{\bf \xi}
$
for $t=1,2,\dots$. Consider the vector of one-step losses
$
{\bf \tilde s_t}={\bf s_t}-{\bf \xi}\left(\frac{1}{\epsilon'_t}-
\frac{1}{\epsilon'_{t-1}}\right)
$
for the moment.

For any vector $\bf s$ and a unit vector $\bf d$ denote
\begin{eqnarray*}
M({\bf s})={\rm argmin}_{{\bf d}\in D}\{{\bf d}\cdot {\bf s}\},
\end{eqnarray*}
where $D=\{(0,\dots 1),\dots ,(1,\dots 0)\}$ is the set of N
unit vectors of dimension N and ``$\cdot$'' is the inner
product of two vectors.

We first show that
\begin{eqnarray}\label{infis-1}
\sum\limits_{t=1}^T M({\bf \tilde s_{1:t}})\cdot{\bf \tilde s_t}\le
M({\bf \tilde s_{1:T}})\cdot\tilde {\bf s_{1:T}}.
\end{eqnarray}
For $T=1$ this is obvious. For the induction step from $T-1$ to $T$
we need to show that
$$
M({\bf \tilde s_{1:T}})\cdot{\bf \tilde s_T}\le
M({\bf \tilde s_{1:T}})\cdot{\bf \tilde s_{1:T}}-
M({\bf \tilde s_{1:T-1}})\cdot {\bf \tilde s_{1:T-1}}.
$$
This follows from ${\bf \tilde s_{1:T}}=
{\bf \tilde s_{1:T-1}}+{\bf \tilde s_T}$ and
$$
M({\bf \tilde s_{1:T}})\cdot{\bf \tilde s_{1:T-1}}\ge
M({\bf \tilde s_{1:T-1}})\cdot{\bf \tilde s_{1:T-1}}.
$$
We rewrite (\ref{infis-1}) as follows
\begin{eqnarray}\label{infis-2}
\sum\limits_{t=1}^T M({\bf \tilde s_{1:t}})\cdot {\bf s_t}\le
M({\bf \tilde s_{1:T}})\cdot{\bf \tilde s_{1:T}}+
\sum\limits_{t=1}^T M({\bf \tilde s_{1:t}})\cdot{\bf \xi}
\left(\frac{1}{\epsilon'_t}-\frac{1}{\epsilon'_{t-1}}\right).
\label{rr-1}
\end{eqnarray}
By definition of $M$ we have
\begin{eqnarray}
M({\bf \tilde s_{1:T}})\cdot{\bf \tilde s_{1:T}}\le
M({\bf s_{1:T}})\cdot\left({\bf s_{1:T}}-\frac{{\bf \xi}}{\epsilon'_T}\right)=
\nonumber
\\
\min_{{\bf d}\in D}\{{\bf d}\cdot{\bf  s_{1:T}}\}-M({\bf s_{1:T}})
\cdot\frac{{\bf \xi}}{\epsilon'_T}~~.
\label{term-1}
\end{eqnarray}
The expectation of the last term in (\ref{term-1}) is equal
to $\frac{1}{\epsilon'_T}=\mu_Tv_T$.

The second term of (\ref{rr-1}) can be rewritten
\begin{eqnarray}
\sum\limits_{t=1}^T M({\bf \tilde s_{1:t}})\cdot{\bf \xi}
\left(\frac{1}{\epsilon'_t}-\frac{1}{\epsilon'_{t-1}}\right)=
\nonumber
\\
\sum\limits_{t=1}^{T} (\mu_tv_t-\mu_{t-1}v_{t-1})
M({\bf \tilde s_{1:t}})\cdot{\bf \xi}~~.
\label{w-1}
\end{eqnarray}

We will use the inequality for mathematical expectation $E$
\begin{eqnarray}
0\le E(M({\bf \tilde s_{1:t}})\cdot{\bf \xi})\le E(M({\bf \xi})\cdot{\bf \xi})=
E(\max_i\xi^i)\le 1+\ln N.
\label{exp-2}
\end{eqnarray}
The proof of this inequality uses ideas of Lemma 1 from~\cite{HuP2004}.

We have for the exponentially distributed random variables
$\xi^i$, $i=1,\dots N$,
\begin{eqnarray}
P\{\max_i\xi^i\ge a\}=P\{\exists i(\xi^i\ge a)\}\le
\sum\limits_{i=1}^N P\{\xi^i\ge a\}=N\exp\{-a\}.
\label{exp-1}
\end{eqnarray}
Since for any non-negative random variable $\eta$,
$
E(\eta)=\int\limits_0^\infty P\{\eta\ge y\}dy,
$
by (\ref{exp-1}) we have
\begin{eqnarray*}
E(\max_i\xi^i-\ln N)=\int\limits_0^\infty P\{\max_i\xi^i-\ln N\ge y\}dy\le
\nonumber
\\
\int\limits_0^\infty N\exp\{-y-\ln N\}dy=1.
\end{eqnarray*}
Therefore, $E(\max_i\xi^i)\le 1+\ln N$.

By (\ref{exp-2}) the expectation of (\ref{w-1}) has the upper bound
\begin{eqnarray*}
\sum\limits_{t=1}^{T}E(M({\bf \tilde s_{1:t}})\cdot{\bf \xi})
(\mu_tv_t-\mu_{t-1}v_{t-1})\le
(1+\ln N)\sum\limits_{t=1}^{T}\mu_t\Delta v_t.
\end{eqnarray*}
Here we have used the inequality $\mu_t\le\mu_{t-1}$ for all $t$,

Since $E(\xi^i)=1$ for all $i$, the expectation of the last
term in (\ref{term-1}) is equal to
\begin{eqnarray}
E\left(M({\bf s_{1:T}})\cdot\frac{\xi}{\epsilon'_T}\right)=
\frac{1}{\epsilon'_T}=\mu_Tv_T.
\label{bou-2}
\end{eqnarray}

Combining the bounds (\ref{infis-2})-(\ref{w-1}) and
(\ref{bou-2}), we obtain
\begin{eqnarray}
r_{1:T}=E\left(\sum\limits_{t=1}^T
M({\bf \tilde s_{1:t}})\cdot{\bf s_t}\right)\le
\nonumber
\\
\min_i s^i_{1:T}-\mu_Tv_T+(1+\ln N)
\sum\limits_{t=1}^{T}\mu_t\Delta v_t\le
\nonumber
\\
\min_i s^i_{1:T}+(1+\ln N)\sum\limits_{t=1}^{T}\mu_t\Delta v_t.
\end{eqnarray}
Lemma is proved. $\triangle$.

We finish now the proof of the theorem.

The inequality (\ref{thh-1a}) of Lemma~\ref{fpl-ifpl-1}
and the inequality (\ref{ii-ff}) of Lemma~\ref{IFPL-1}
imply the inequality
\begin{eqnarray}
E(s_{1:T})\le\min_i s^i_{1:T}+
\nonumber
\\
+\sum_{t=1}^T (2(e^{3/a}-1)(\gamma(t))^{1-\alpha_t}+a(1+\ln N)
(\gamma(t))^{\alpha_t})\Delta v_t.
\label{j_mu-1af}
\end{eqnarray}
for all $T$.

The optimal value (\ref{optalpha-1}) of $\alpha_t$ can be
easily obtained by minimization of each member of the sum
(\ref{j_mu-1af}) by $\alpha_t$.
In this case $\mu_t$ is equal to (\ref{mu-1}) and
(\ref{j_mu-1af}) is equivalent to
\begin{eqnarray}
E(s_{1:T})\le\min_i s^i_{1:T}+2\sqrt{2a(e^{3/a}-1)(1+\ln N)}
\sum_{t=1}^T (\gamma(t))^{1/2}\Delta v_t,
\label{main-refret-1j}
\end{eqnarray}
where $a$ is a parameter of the algorithm {\rm PROT}.

Also, for each $\epsilon>0$ an
$a$ exists such that $2a(e^{3/a}-1)<6+\epsilon$. Therefore, we obtain
(\ref{main-refret-1}).

We have $\sum_{t=1}^T \Delta v_t=v_T$ for all $T$,
$v_t\to\infty$ and $\gamma(t)\to 0$ as $t\to\infty$. Then by
Toeplitz lemma (see Lemma~\ref{toep-1} of
Section~\ref{lem-11p})
\begin{eqnarray*}
\frac{1}{v_T}\left(2\sqrt{(6+\epsilon)(1+\ln N)}
\sum_{t=1}^T (\gamma(t))^{1/2}\Delta v_t\right)\to 0
\label{j_mu-1ass}
\end{eqnarray*}
as $T\to\infty$. Therefore, the {\rm FPL} algorithm {\rm PROT}
is asymptotically consistent in the mean, i.e., the relation
(\ref{mod-consist-1hhh}) of Theorem~\ref{cor-2} is proved.
$\triangle$

In case where all losses are nonnegative: $s_t^i\in [0,+\infty)$,
the inequality (\ref{ij-1}) can be replaced on
\begin{eqnarray*}\label{ij-1a}
\left|\frac{s_{1:t-1}^j-s^{j_2}_{1:t-1}}{v_{t-1}}\right|\le 1
\end{eqnarray*}
for all $t$ and $i$. In this case an analysis of the proof of
Lemma~\ref{fpl-ifpl-1}
shows that the bound (\ref{main-refret-1j}) can be replaced on
\begin{eqnarray*}
E(s_{1:T})\le\min_i s^i_{1:T}+2\sqrt{a(e^{2/a}-1)(1+\ln N)}
\sum_{t=1}^T (\gamma(t))^{1/2}\Delta v_t,
\end{eqnarray*}
where $a$ is a parameter of the algorithm {\rm PROT}.

Since for each $\epsilon>0$ an
$a$ exists such that $a(e^{2/a}-1)<2+\epsilon$, we obtain
a version of (\ref{main-refret-1}) for nonnegative losses -- the inequality
(\ref{main-refret-1kk}).

We study now the Hannan consistency of our algorithm.

\begin{theorem}\label{cor-2s}
Assume that all conditions of Theorem~\ref{cor-2s} hold and
\begin{eqnarray}
\sum\limits_{t=1}^\infty (\gamma(t))^2<\infty.
\label{hhann-1}
\end{eqnarray}
Then the algorithm {\rm PROT} is Hannan consistent:
\begin{eqnarray}
\limsup\limits_{T\to\infty}
\frac{1}{v_T}\left(s_{1:T}-\min\limits_{i=1,\dots N} s^i_{1:T}\right)\le 0
\label{consist-1han-1h}
\end{eqnarray}
almost surely.
\end{theorem}
{\it Proof.} So far we assumed that perturbations $\xi^{1},
\dots ,\xi^{N}$ are sampled only once at time $t=0$. This
choice was favorable for the analysis. As it easily seen, under
expectation this is equivalent to generating new perturbations
$\xi_{t}^{1}, \dots ,\xi_{t}^{N}$ at each time step $t$; also,
we assume that all these perturbations are i.i.d for $i=1,\dots
,N$ and $t=1,2,\dots$. Lemmas~\ref{fpl-ifpl-1}, \ref{IFPL-1}
and Theorem~\ref{cor-2} remain valid for this case. This method
of perturbation is needed to prove the Hannan consistency of
the algorithm {\rm PROT}.

We use some version of the strong law of large numbers to prove
the Hannan consistency of the algorithm {\rm PROT}.
\begin{proposition}\label{lemma-2}
Let $g(x)$ be a positive nondecreasing real function
such that $x/g(x)$, $g(x)/x^2$ are non-increasing for $x>0$ and $g(x)=g(-x)$
for all $x$.

Let the assumptions of Theorem~\ref{cor-2} hold and
\begin{eqnarray}
\sum\limits_{t=1}^\infty\frac{g(\Delta v_t)}{g(v_t)}<\infty.
\label{gg-1}
\end{eqnarray}
Then the {\rm FPL} algorithm {\rm PROT} is Hannan consistent, i.e.,
(\ref{consist-1han-1}) holds as $T\to\infty$ almost surely.
\end{proposition}
{\it Proof.} The proof is based on the following lemma.

\begin{lemma}\label{lem-11}
Let $a_t$ be a nondecreasing sequence of real numbers such that
$a_t\to\infty$ as $t\to\infty$ and $X_t$ be a sequence of
independent random variables such that $E(X_t)=0$, for
$t=1,2,\dots$. Let also, $g(x)$ satisfies assumptions of
Proposition~\ref{lemma-2}. Then the inequality
\begin{eqnarray}
\sum\limits_{t=1}^\infty\frac{E(g(X_t))}{g(a_t)}<\infty
\label{suff-law-1}
\end{eqnarray}
implies
\begin{eqnarray}
\frac{1}{a_T}\sum\limits_{t=1}^T X_t\to 0
\label{law-12}
\end{eqnarray}
as $T\to\infty$ almost surely.
\end{lemma}
The proof of this lemma is given in Section~\ref{lem-11p}.

Put $X_t=(s_t-E(s_t))/2$, where $s_t$ is the loss of the {\rm FPL}
algorithm {\rm PROT} at step $t$, and $a_t=v_t$ for all $t$.
By definition $|X_t|\le\Delta v_t$ for all $t$.
Then (\ref{suff-law-1}) is valid, and by (\ref{law-12})
$$
\frac{1}{v_T}(s_{1:T}-E(s_{1:T}))=
\frac{1}{v_T}\sum\limits_{t=1}^T (s_t-E(s_t))\to 0
$$
as $T\to\infty$ almost surely.
This limit and the limit (\ref{mod-consist-1hhh}) imply (\ref{consist-1han-1h}).
$\triangle$

By Lemma~\ref{lemma-2} the algorithm {\rm PROT} is Hannan
consistent, since (\ref{hhann-1}) implies (\ref{gg-1}) for
$g(x)=x^2$. Theorem~\ref{cor-2s} is proved. $\triangle$

Authors of~\cite{AAAGO2006} and~\cite{PoHP2005} considered
polynomially bounded one-step losses. We consider a specific
example of the bound (\ref{j_mu-1af}) for polynomial case.
\begin{corollary}\label{cor-1gg}
Assume that $|s^i_t|\le t^\alpha$ for all $t$ and $i=1,\dots N$,
and $v_t\ge t^{\alpha+\delta}$ for all $t$, where $\alpha$ and
$\delta$ are positive real numbers. Let also, in the algorithm
{\rm PROT}, $\gamma(t)=t^{-\delta}$ and
$\mu_t=a(\gamma(t))^{\alpha_t}$, where $\alpha_t$ is defined by
(\ref{optalpha-1}). Then
\begin{itemize}
\item{(i)} the algorithm {\rm PROT} is asymptotically
    consistent in the mean for any $\alpha>0$ and
    $\delta>0$;
\item{(ii)} this algorithm is Hannan consistent for any
    $\alpha>0$ and $\delta>\frac{1}{2}$;
\item{(iii)}
the expected loss of this algorithm is bounded~:
\begin{eqnarray}
E(s_{1:T})\le\min_i s^i_{1:T}+2\sqrt{(6+\epsilon)(1+\ln N)}T^{1-\frac{1}{2}\delta+\alpha}
\label{poly-1}
\end{eqnarray}
as $T\to\infty$, where $\epsilon>0$ is a parameter of the
algorithm.\footnote{Recall that given $\epsilon$ we tune
the parameter $a$ of the algorithm {\rm PROT}. }
\end{itemize}
\end{corollary}
This corollary follows directly from Theorem~\ref{cor-2}, where
condition~(\ref{hhann-1}) of Theorem~\ref{cor-2}
holds for $\delta>\frac{1}{2}$.

If $\delta=1$ the regret from (\ref{poly-1}) is asymptotically equivalent to
the regret from Allenberg et al.~\cite{AAAGO2006} (see Section~\ref{intr-1}).

For $\alpha=0$ we have the case of bounded loss function ($|s^i_t|\le 1$ 
for all $i$ and $t$). The {\rm FPL} algorithm {\rm
PROT} is asymptotically consistent in the mean if
$v_t\ge\beta(t)$ for all $t$, where $\beta(t)$ is an arbitrary
positive unbounded non-decreasing computable function (we can
get $\gamma(t)=1/\beta(t)$ in this case). This algorithm is
Hannan consistent if (\ref{hhann-1}) holds, i.e.
$$
\sum\limits_{t=1}^\infty(\beta(t))^{-2}<\infty.
$$
For example, this condition be satisfied for $\beta(t)=t^{1/2}\ln t$.

Theorem~\ref{cor-2} is also valid for the standard time
scaling, i.e., when $v_T=T$ for all $T$, and when losses of
experts are bounded, i.e., $\alpha=0$. Then for any $\epsilon>0$ the
expected regret has the upper bound
$$
2\sqrt{(6+\epsilon)(1+\ln N)}\sum_{t=1}^T (\gamma(t))^{1/2}\le
4\sqrt{(6+\epsilon)(1+\ln N)T}
$$
which is similar to bounds from \cite{HuP2004} and \cite{KaV2003}.


Let us show that the bound (\ref{main-refret-1}) of
Theorem~\ref{cor-2} that holds against oblivious experts also
holds against non-oblivious (adaptive) ones.

In non-oblivious case, it is natural to generate at each time
step $t$ of the algorithm {\rm PROT} a new vector of
perturbations ${\bf \bar\xi_{t}}=(\xi_{t}^{1}, \dots ,\xi_{t}^{N})$,
$\bar\xi_{0}$ is empty set. Also, it is assumed that all these
perturbations are i.i.d according to the exponential
distribution $P$, where $i=1,\dots ,N$ and $t=1,2,\dots$.
Denote ${\bf \bar\xi_{1:t}}=(\bar\xi_{1},\dots ,\bar\xi_{t})$.

Non-oblivious experts can react at each time step $t$ on past
decisions $s_{1},s_{2},\dots s_{t-1}$ of the {\rm FPL}
algorithm and on values of $\bar\xi_{1},\dots ,\bar\xi_{t-1}$.

Therefore, losses of experts and regret depend now from random
perturbations:
\begin{eqnarray*}
s^{i}_{t}=s^{i}_{t}({\bf \bar\xi_{1:t-1}}), \mbox{ }i=1,\dots ,N,
\nonumber
\\
\Delta v_{t}=\Delta v_{t}({\bf \bar\xi_{1:t-1}}),
\end{eqnarray*}
where $t=1,2,\dots$.

In non-oblivious case, condition (\ref{bound-fluc-1}) is a random
event. We assume in Theorem~\ref{cor-2} that in the game of prediction
with expert advice regulated by the {\rm FPL}-protocol the event
\begin{eqnarray*}\label{bound-fluc-1g}
\Fluc(t)\le\gamma(t)\mbox{ for all }t
\end{eqnarray*}
holds almost surely.

An analysis of the proof of Theorem~\ref{cor-2} shows that in
non-oblivious case, the bound (\ref{main-refret-1}) is an
inequality for the random variable
\begin{eqnarray}
\sum_{t=1}^T E(s_{t})-\min_i s^i_{1:T}-
\nonumber
\\
-2\sqrt{(6+\epsilon)(1+\ln N)} \sum_{t=1}^T
(\gamma(t))^{1/2}\Delta v_t\le 0, \label{main-refret-2a}
\end{eqnarray}
which holds almost surely with respect to the product
distribution $P^{t-1}$, where the loss of the {\rm FPL}
algorithm $s_{t}$ depend on a random perturbation $\xi_{t}$ at
step $t$ and on losses of all experts on steps $<t$. Also, $E$
is the expectation with respect to $P$.

Taking expectation $E_{1:T-1}$ with respect to the product
distribution $P^{t-1}$ we obtain a version of
(\ref{main-refret-1}) for non-oblivious case
\begin{eqnarray*}
E_{1:T}\left(s_{1:T}-\min_i s^i_{1:T}-
2\sqrt{(6+\epsilon)(1+\ln N)}
\sum_{t=1}^T (\gamma(t))^{1/2}\Delta v_t\right)\le 0
\end{eqnarray*}
for all $T$.

\section{An example: zero-sum experts}\label{arbitr-1}

In this section we present an example of a game, where losses
of experts cannot be bounded~\cite{Vyu2009a} in advance. Let
$S=S(t)$ be a function representing evolution of a stock price.
Two experts will represent two concurrent methods of buying and
selling shares of this stock.

\begin{figure}[t]
\centering\includegraphics[height=100mm,width=150mm,clip]{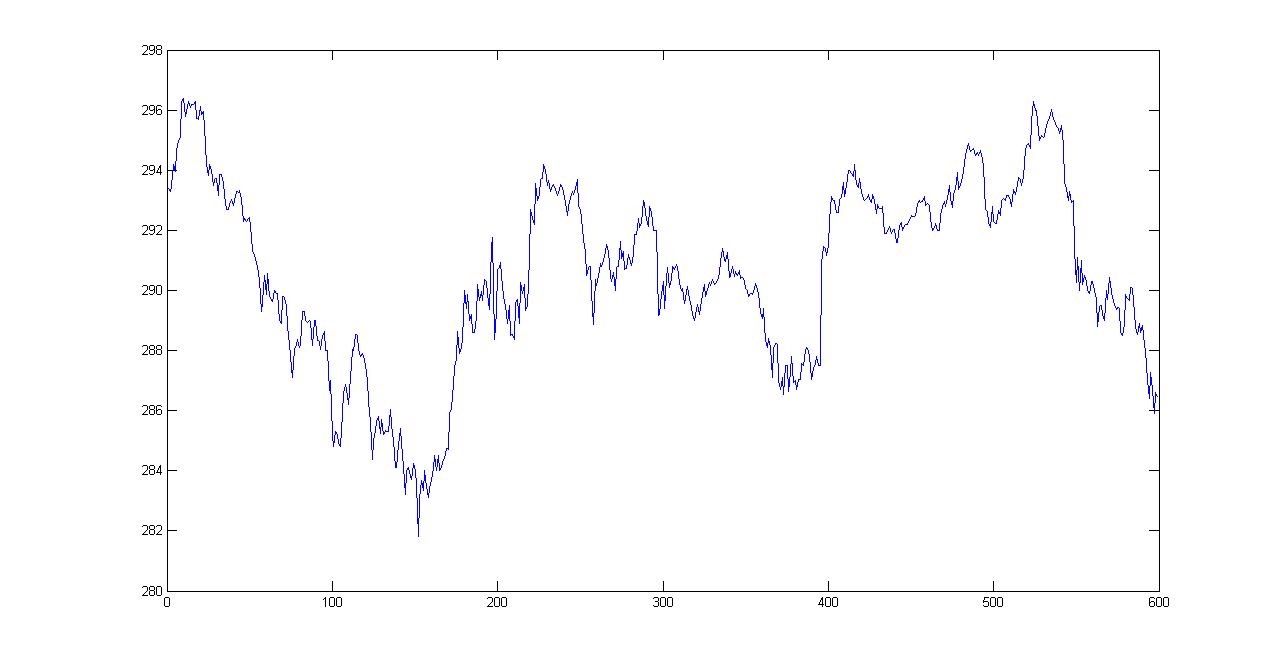}
         \begin{center}
         {\small Fig. 1. Evolution of a stock price}
         \end{center}
                  \end{figure}

\begin{figure}[t]
\centering\includegraphics[height=100mm,width=150mm,clip]{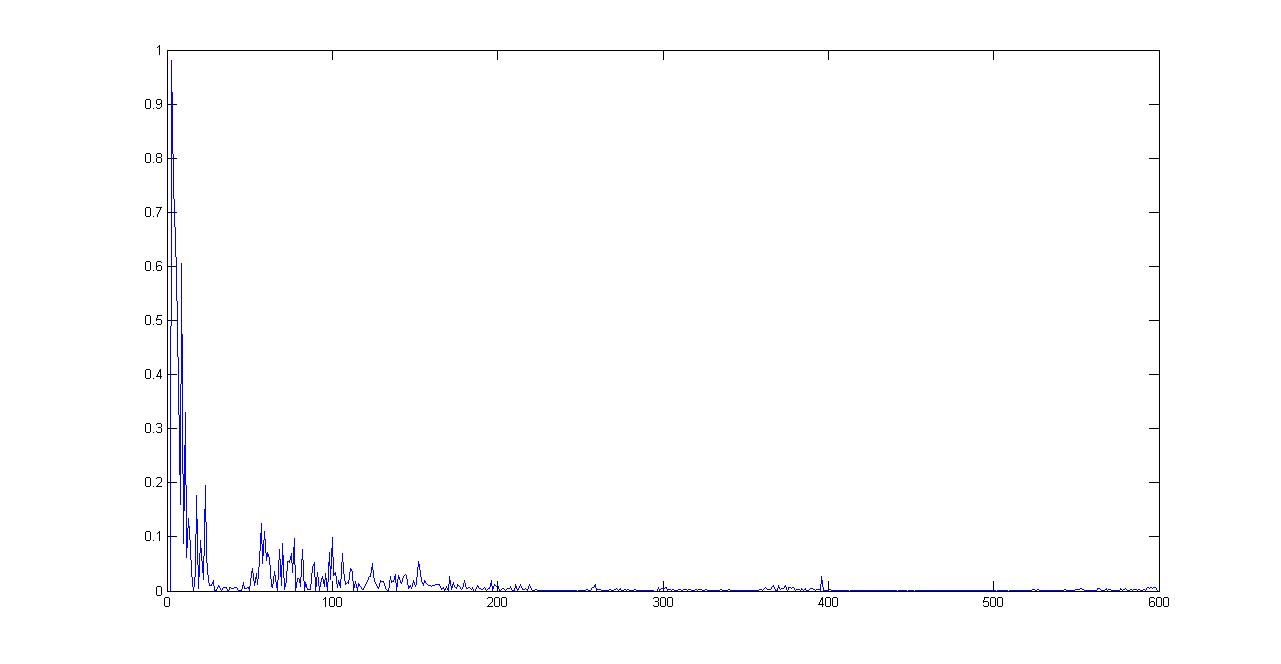}
         \begin{center}
         {\small Fig. 2. Fluctuation of the game}
         \end{center}
                  \end{figure}

\begin{figure}[t]
\centering\includegraphics[height=100mm,width=150mm,clip]{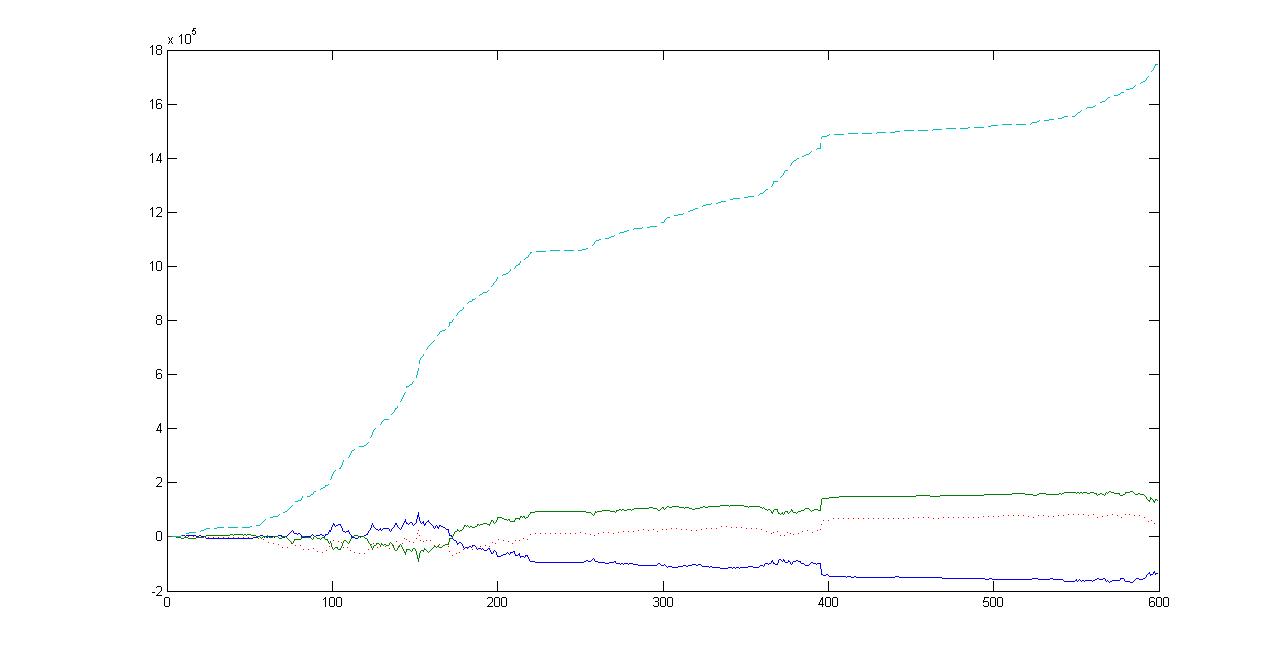}
         \begin{center}
         {\small Fig. 3. Two symmetric solid lines -- gains of two zero
sums strategies,
         dotted line -- expected gain of the algorithm {\rm PROT},
dashed line -- volume of the game}
         \end{center}
                  \end{figure}

Let $M$ and $T$ be positive integer numbers and let the time
interval $[0,T]$ be divided on a large number $M$ of
subintervals. Define a discrete time series of stock prices
\begin{eqnarray}
S_0=S(0), S_1=S(T/(M)),S_2=S(2T/(M))\dots , S_{M}=S(T).
\label{time-1}
\end{eqnarray}
In this paper, volatility is an informal notion. We say that
the difference $(S_{T}-S_{0})^2$ represents the macro
volatility and the sum $\sum\limits_{i=0}^{T-1}(\Delta S_i)^2$,
where $\Delta S_i=S_{i+1}-S_i$, $i=1,\dots T-1$, represents the
micro volatility of the time series (\ref{time-1}).

The game between an investor and the market looks as follows:
the investor can use the long and short selling. At beginning
of time step $t$ Investor purchases the number $C_t$ of shares
of the stock by $S_{t-1}$ each. At the end of trading period
the market discloses the price $S_{t+1}$ of the stock, and the
investor incur his current income or loss $s_t=C_t\Delta S_t$
at the period $t$. We have the following equality
\begin{eqnarray}
(S_T-S_0)^2=(\sum\limits_{t=0}^{T-1}\Delta S_t)^2=
\nonumber
\\
=\sum\limits_{t=0}^{T-1}
2(S_t-S_0)\Delta S_t+\sum\limits_{t=0}^{T-1}(\Delta S_t)^2.
\label{main-1}
\end{eqnarray}
The equality (\ref{main-1}) leads to the two strategies for
investor which are represented by two experts. At the beginning
of step $t$ Experts 1 and 2 hold the number of shares
\begin{eqnarray}
C^1_t=2C(S_t-S_0),
\label{stat-1}
\\
C^2_t=-C^1_t,
\label{stat-2}
\end{eqnarray}
where $C$ is an arbitrary positive constant.

These strategies at step $t$ earn the incomes
$s^1_t=2C(S_t-S_0)\Delta S_t$ and $s^2_t=-s^1_t$. The strategy
(\ref{stat-1}) earns in first $T$ steps of the game the income
$$
s^1_{1:T}=\sum\limits_{t=1}^T s^1_t=
2C((S_T-S_0)^2-\sum\limits_{t=1}^{T-1}(\Delta S_t)^2).
$$
The strategy (\ref{stat-2}) earns in first $T$ steps the income
$s^2_{1:T}=-s^1_{1:T}$.

The number of shares $C^1_t$ in the strategy (\ref{stat-1}) or
number of shares $C^2_t=-C^1_t$ in the strategy (\ref{stat-2})
can be positive or negative. The one-step gains $s^1_t$ and
$s^2_t=-s^1_t$ are unbounded and can be positive or negative:
$s^i_t\in (-\infty,+\infty)$.


Informally speaking, the first strategy will show a large
return if $(S_T-S_0)^2\gg\sum\limits_{i=0}^{T-1}(\Delta
S_i)^2$; the second one will show a large return when
$(S_T-S_0)^2\ll\sum\limits_{i=0}^{T-1}(\Delta S_i)^2$. There is
an uncertainty domain for these strategies, i.e., the case when
both $\gg$ and $\ll$ do not hold. The idea of these strategies
is based on the paper of Cheredito~\cite{che} (see also
Rogers~\cite{Rog}, Delbaen and Schachermayer~\cite{DeS}) who
have constructed arbitrage strategies for a financial market
that consists of money market account and a stock whose price
follows a fractional Brownian motion with drift or an
exponential fractional Brownian motion with drift.
Vovk~\cite{Vov} has reformulated these strategies for discrete
time. We use these strategies to define a mixed strategy which
incur gain when macro and micro volatilities of time series
differ. There is no uncertainty domain for continuous time.


We analyze this game in the decision theoretic online learning
(DTOL) framework~\cite{FrS97}. We introduce {\it Learner} that
can choose between two strategies (\ref{stat-1}) and
(\ref{stat-2}). To change from the follow leader framework to
DTOL we derandomize the {\rm FPL} algorithm {\rm
PROT}.\footnote{To apply Theorem~\ref{cor-2} we interpreted
gain as a negative loss.} We interpret the expected one-step
gain $E(s_t)$ gain as the weighted average of one-step gains of
experts strategies. In more detail, at each step $t$, {\it
Learner} divide his investment in proportion to the
probabilities of expert strategies (\ref{stat-1}) and
(\ref{stat-2}) computed by the {\rm FPL} algorithm and suffers
the gain
\begin{eqnarray*}
G_t=2C(S_t-S_0)(P\{I_t=1\}-P\{I_t=2\})\Delta S_t
\end{eqnarray*}
at any step $t$, where $C$ is an arbitrary positive constant;
$G_{1:T}=\sum_{t=1}^T G_t=E(s_{1:T})$ is the {\it Learner's}
cumulative gain.

Assume that $|s^1_t|=o(\sum_{i=1}^t|s^1_i|)$ as $t\to\infty$.
Let $\gamma(t)=\mu$ for all $t$, where $\mu$ is
arbitrary small positive number. Then for any $\epsilon>0$
$$
G_{1:T}\ge\left|\sum_{t=1}^T s^1_t\right|-2\mu^{1/2}\sqrt{(6+\epsilon)(1+\ln N)}
\left(\sum_{t=1}^T|s^1_t|+v_{0}\right)
$$
for all sufficiently large $T$, and for some $v_{0}\ge 0$.

Under condition of Theorem~\ref{cor-2} we show that strategy of algorithm {\rm PROT} is
``defensive'' in some weak sense~:
$$
G_{1:T}-\left|\sum_{t=1}^T s^1_t\right|\ge
-o\left(\sum_{t=1}^T|s^1_t|+v_{0}\right)
$$
as $T\to\infty$.

\section{Conclusion}

In this paper we try to extend methods of the theory of
prediction with expert advice for the case when experts
one-step gains cannot be bounded in advance. The traditional
measures of performance do not work in general unbounded
case. To measure the asymptotic performance of our algorithm, we
replace the traditional time-scale on a volume-scale. New
notion of volume of a game and scaled fluctuation of a game
are introduced in this paper. In case
of two zero-sum experts this notion corresponds to the sum of all
transactions between experts.

Using the notion of the scaled fluctuation of a game, we can define
very broad classes of games (experts) for which our algorithm
{\rm PROT} is asymptotically consistent in the modified sense.
Also, restrictions on
such games are formulated in relative terms: the logarithmic
derivative of the volume of the game must be $o(t)$ as
$t\to\infty$.

A motivating example of a game with two zero-sum
experts from Section~\ref{arbitr-1} shows some practical
significance of these problem. The {\rm FPL} algorithm with
variable learning rates is simple to implement and it is
bringing satisfactory experimental results when prices follow
fractional Brownian motion.

There are some open problems for further research. It would be
useful to analyze the performance of the well known algorithms
from DTOL framework (like ``Hedge''~\cite{FrS97} or ``Normal
Hedge''~\cite{CFH2009}) for the case of unbounded losses in
terms of the volume of a game.

There is a gap between Proposition~\ref{cor-2-1} and
Theorem~\ref{cor-2}, since we assume in this theorem that the
game satisfies $\Fluc(t)\le\gamma(t)\to 0$, where $\gamma(t)$
is computable. Also, the function $\gamma(t)$ is a parameter of
our algorithm {\rm PROT}. Does there exists an asymptotically
consistent learning algorithm in case where $\Fluc(t)\to 0$ as
$t\to\infty$ and where the function $\gamma(t)$ is not a
parameter of this algorithm?

A partial solution is based on applying ``double trick'' method
to an increasing sequence of nonnegative functions
$\gamma_{i}(t)$ such that $\gamma_{i}(t)\to 0$ as $t\to\infty$
and $\gamma_{i}(t)\le \gamma_{i+1}(t)$ for all $i$ and $t$. In
this case a modified algorithm {\rm PROT} is asymptotically
consistent in the mean in any game such that
$$\limsup\limits_{t\to\infty}\frac{\Fluc(t)}{\gamma_{i}(t)}<\infty$$
for some $i$.

We consider in this paper only the full information case. An
analysis of these problems under partial monitoring is a
subject for a further research.



\appendix
\section{Proof of Lemma~\ref{lem-11}}\label{lem-11p}

The proof of Lemma~\ref{lem-11} is based on Kolmogorov's
theorem on three series and its corollaries. For completeness
of presentation we reconstruct the proof from
Petrov~\cite{Pet75}~(Chapter IX, Section 2).

For any random variable $X$ and a positive number $c$ denote
\[
X^{c}=
  \left\{
    \begin{array}{l}
      X \mbox{ if } |X|\le c
    \\
      0 \mbox{ otherwise. }
    \end{array}
  \right.
\]
The Kolmogorov's theorem on three series says:

For any sequence of independent random variables $X_{t}$,
$t=1,2,\dots$, the following implications hold
\begin{itemize}
\item{} If the series $\sum_{t=1}^{\infty}X_{t}$ is
    convergent almost surely then the series
    $\sum_{t=1}^{\infty}EX^{c}_{t}$,
    $\sum_{t=1}^{\infty}DX^{c}_{t}$ and
    $\sum_{t=1}^{\infty}P\{|X_{t}|\ge c\}$ are convergent
    for each $c>0$, where $E$ is the mathematical
    expectation and $D$ is the variation.
\item{} The series $\sum_{t=1}^{\infty}X_{t}$ is convergent
    almost surely if all these series are convergent for
    some $c>0$.
\end{itemize}
See Shiryaev~\cite{Shi80} for the proof.

Assume conditions of Lemma~\ref{lem-11} hold. We will prove
that
\begin{eqnarray}
\sum\limits_{t=1}^{\infty}\frac{Eg(X_{t})}{g(a_{t})}<\infty
\label{g-t-1}
\end{eqnarray}
implies
\begin{eqnarray}
\sum\limits_{t=1}^{\infty}\frac{X_{t}}{a_{t}}<\infty
\nonumber
\end{eqnarray}
almost surely. From this, by Kroneker's lemma~\ref{kron-1} (see
below), the series
\begin{eqnarray}
\frac{1}{a_{t}}\sum\limits_{t=1}^{\infty}X_{t}
\label{corr-res-1}
\end{eqnarray}
is convergent almost surely.

Let $V_{t}$ be a distribution function of the random variable
$X_{t}$. Since $g$ non-increases,
\begin{eqnarray*}
P\{|X_{t}|>a_{t}\}\le\int_{|x|\ge a_{t}}\frac{g(x)}{g(a_{t})}dV_{t}(x)\le
\frac{Eg(X_{t})}{g(a_{t})}.
\end{eqnarray*}
Then by (\ref{g-t-1})
\begin{eqnarray}
\sum\limits_{t=1}^{\infty}P\left\{\left|\frac{X_{t}}{a_{t}}\right|\ge 1\right\}<\infty
\label{first-1}
\end{eqnarray}
almost surely. Denote
\[
Z_{t}=
  \left\{
    \begin{array}{l}
      X_{t} \mbox{ if } |X_{t}|\le a_{t}
    \\
      0 \mbox{ otherwise. }
    \end{array}
  \right.
\]
By definition $x^{2}/g(x))\le a_{t}/g(a_{t})$ for $|x|<a_{t}$.
Rearranging, we obtain $x^{2}/a_{t}\le g(x)/g(a_{t})$ for these
$x$. Therefore,
\begin{eqnarray*}
EZ_{t}^{2}=\int\limits_{|x|<a_{t}}x^{2} dV_{t}(x)\le
\frac{a_{t}^{2}}{g(a_{t})}\int\limits_{|x|<a_{t}}g(x)dV_{t}(x)\le
\frac{a_{t}^{2}}{g(a_{t})}Eg(X_{t}).
\end{eqnarray*}
By (\ref{g-t-1}) we obtain
\begin{eqnarray}
\sum\limits_{t=1}^{\infty}E\left(\frac{Z_{t}}{a_{t}}\right)^{2}<\infty.
\label{g-t-2}
\end{eqnarray}
Since $EX_{t}=\int\limits_{-\infty}^{\infty} x dV_{t}(x)=0$,
\begin{eqnarray}
|EZ_{t}|=\left|\int\limits_{|x|>a_{t}}xdV_{t}(x)\right|\le
\frac{a_{t}}{g(a_{t})}\int\limits_{|x|>a_{t}}g(x)dV_{t}(x)\le
\frac{a_{t}}{g(a_{t})}Eg(X_{t}).
\label{g-t-3}
\end{eqnarray}
By (\ref{g-t-1})
$$
\sum\limits_{t=1}^{\infty}E\left(\frac{X_{t}}{a_{t}}\right)^{1}\le
\sum\limits_{t=1}^{\infty}\left|E\left(\frac{Z_{t}}{a_{t}}\right)\right|<\infty.
$$
From (\ref{first-1})--(\ref{g-t-3}) and the theorem on three
series we obtain (\ref{corr-res-1}).

We have used Toeplitz and Kroneker's lemmas.
\begin{lemma}\label{toep-1} (Toeplitz)
Let $x_{t}$ be a sequence of real numbers and $b_{t}$ be a
sequence of nonnegative real numbers such that
$a_{t}=\sum\limits_{i=1}^{t}b_{i}\to\infty$, $x_{t}\to x$ and
$|x|<\infty$. Then
\begin{eqnarray}
\frac{1}{a_{t}}\sum\limits_{i=1}^{t}b_{i}x_{i}\to x.
\label{toepli-1}
\end{eqnarray}
\end{lemma}
{\it Proof.} For any $\epsilon>0$ an $t_{\epsilon}$ exists such
that $|x_{t}-x|<\epsilon$ for all $t\ge t_{\epsilon}$. Then
$$
\left|\frac{1}{a_{t}}\sum\limits_{I=1}^{t}b_{i}(x_{i}-x)\right|\le
\frac{1}{a_{t}}\sum\limits_{i<t_{\epsilon}}|b_{i}(x_{i}-x)|+\epsilon
$$
for all $t\ge t_{\epsilon}$. Since $a_{t}\to\infty$, we obtain
(\ref{toepli-1}).
\begin{lemma}\label{kron-1} (Kroneker)
Assume $\sum\limits_{t=1}^{\infty}x_{t}<\infty$ and
$a_{t}\to\infty$ Then
$\frac{1}{a_{t}}\sum\limits_{i=1}^{t}a_{i}x_{i}\to 0$.
\end{lemma}
The proof is the straightforward corollary of Toeplitz lemma.

\end{document}